\documentclass[runningheads]{llncs}
\usepackage[T1]{fontenc}
\usepackage{amsmath}
\usepackage{booktabs}
\usepackage{multirow}
\usepackage[table]{xcolor}
\usepackage{ragged2e}
\usepackage{esvect}
\usepackage{xcolor} 
\definecolor{softblue}{RGB}{60,130,250} 
\usepackage[colorlinks=true,
            linkcolor=black,   
            citecolor=softblue,   
            urlcolor=black]     
            {hyperref}

\usepackage{graphicx}
\begin{document}
\title{PROPEX-RAG: Enhanced GraphRAG using Prompt-Driven Prompt Execution}

\titlerunning{PROPEX-RAG: Enhanced GraphRAG}
%
%
\author{Tejas Sarnaik\orcidID{0009-0009-6834-3847} \and
Manan Shah\orcidID{0009-0003-7996-4768} \and
Ravi Hegde\orcidID{0000-0002-0418-5861}}
\authorrunning{Tejas Sarnaik et al.}
%
\institute{Indian Institute of Technology, Gandhinagar, Gujarat, India
\\
\email{\{tejas.sarnaik, mananshah, hegder\}@iitgn.ac.in}}

\maketitle              
\begin{abstract}
\justifying
Retrieval-Augmented Generation (RAG) has become a robust framework for enhancing Large Language Models (LLMs) with external knowledge. Recent advances in RAG have investigated graph-based retrieval for intricate reasoning; however, the influence of prompt design on enhancing the retrieval and reasoning process is still considerably under-examined. In this paper, we present a prompt-driven GraphRAG framework that underscores the significance of prompt formulation in facilitating entity extraction, fact selection, and passage re-ranking for multi-hop question answering. Our approach creates a symbolic knowledge graph from text data by encoding entities and factual relationships as structured facts triples. We use LLMs selectively during online retrieval to perform semantic filtering and answer generation. We also use entity-guided graph traversal through Personalized PageRank (PPR) to support efficient, scalable retrieval based on the knowledge graph we built. Our system gets \textbf{state-of-the-art performance} on HotpotQA and 2WikiMultiHopQA, with F1 scores of \textbf{80.7\%} and \textbf{78.9\%}, and Recall@5 scores of \textbf{97.1\%} and \textbf{98.1\%}, respectively. These results show that prompt design is an important part of improving retrieval accuracy and response quality. This research lays the groundwork for more efficient and comprehensible multi-hop question-answering systems, highlighting the importance of prompt-aware graph reasoning. Code and data are available at \href{https://github.com/tejas-sarnaik/ProPEX-RAG}{https://github.com/tejas-sarnaik/ProPEX-RAG}.

\end{abstract}
\section{Introduction}
\noindent
\justifying
\sloppy
Large language models (LLMs) have shown significant prowess in understanding and generating natural language, matching human-level performance in varied tasks. However, they face core shortcomings in accessing current information, reasoning with external knowledge, and providing factually verifiable answers, especially in multi-hop tasks that require the synthesis of information from various sources through intricate reasoning chains. The RAG paradigm improves LLM by enabling dynamic access to external knowledge bases. Traditional RAG uses dense retrieval to identify relevant passages, conditioned on the contexts retrieved \cite{karpukhin2020dense}. This method is effective for single-hop queries, but it often fails in multihop reasoning where the relevant data is distributed across multiple distinct passages \cite{xiong2021answering}. New methods explore graph-based retrieval for advanced reasoning. Graph structures illustrate explicit entity relationships, supporting inferential pathways similar to human reasoning. HippoRAG2 used neurobiological architectures to improve performance by up to 20\%, with cost reductions of 10-30x \cite{gutierrez2025rag}. These innovations underscore graph-based strategies as powerful for complex tasks requiring integrated knowledge sourcing. Despite progress, the role of prompt design in RAG system efficiency is under explored. Current methods treat it as a secondary optimization rather than a central feature of the architecture. This neglect is critical given the evidence that prompt design affects entity extraction accuracy, fact selection , and passage classification. Step-back prompting alone increases the precision of complex reasoning when used with retrieval. This research presents a prompt-based GraphRAG framework prioritizing prompt design to enhance retrieval-augmented generation, featuring prompt-driven execution integration, embedding prompt engineering throughout the pipeline, and prompt-guided semantic filtering, using prompts for precise sub-graph selections, maintaining semantic integrity, and reducing computational demands.

\subsubsection{Related Work:}
\noindent
\justifying
\sloppy
Recent advancements in retrieval-augmented generation propose that incorporating symbolic structures, notably knowledge graphs, into retrieval can substantially enhance outcomes. Traditionally, RAG employs dense vector retrieval and semantic similarity to support assertions, which often fails to uncover link structures vital for advanced reasoning. Graph-Based RAG solves this by integrating knowledge graph connections, simplifying discovery. NodeRAG \cite{xu2025noderag}, an RAG variant, uses various graph nodes that show interconnections in long-form queries, thus optimizing multi-hop question answering (QA). Document GraphRAG helps to understand documents using knowledge graphs \cite{knollmeyer2025document}, improving search efficiency across multiple texts. KG-Based RAG employs graph vector embeddings to enhance retrieval semantics \cite{shavaki2024knowledge}, vital for multi-hop QA requiring interconnected evidence synthesis and data flow tracking. Current research transcends linear search, inspired by datasets like HotpotQA and 2WikiMultiHopQA. Fact-Centric Knowledge Web \cite{sinha2024factcentric} highlight fact-centric webs for domain-agnostic retrieval, focusing on prompt-sensitive symbolic reasoning. Query-Aware GNN \cite{agrawal2025queryaware} enhance retrieval using query-sensitive graph neural networks, demonstrating the role of question structure in graph-based reasoning for complex multi-hop inference.
Beyond graph-augmented retrieval, recent work elevates prompt policies to first-class, auditable components that steer seeding, local fact filtering, typed graph traversal, and evidence presentation at test time \cite{wei2022chain}. Rather than re-training large retrievers or relying on heavy learned re-rankers, a compact policy layer instantiated via precise prompts and light controllers enables controllable navigation in the spirit of topic-sensitive PageRank, transparent error surfaces, and low-friction domain adaptation \cite{haveliwala2002topic,peng2024graphrag_survey}.

\section{Proposed Methodology}
\begin{figure}[h]
\centering
\includegraphics[width=\textwidth]{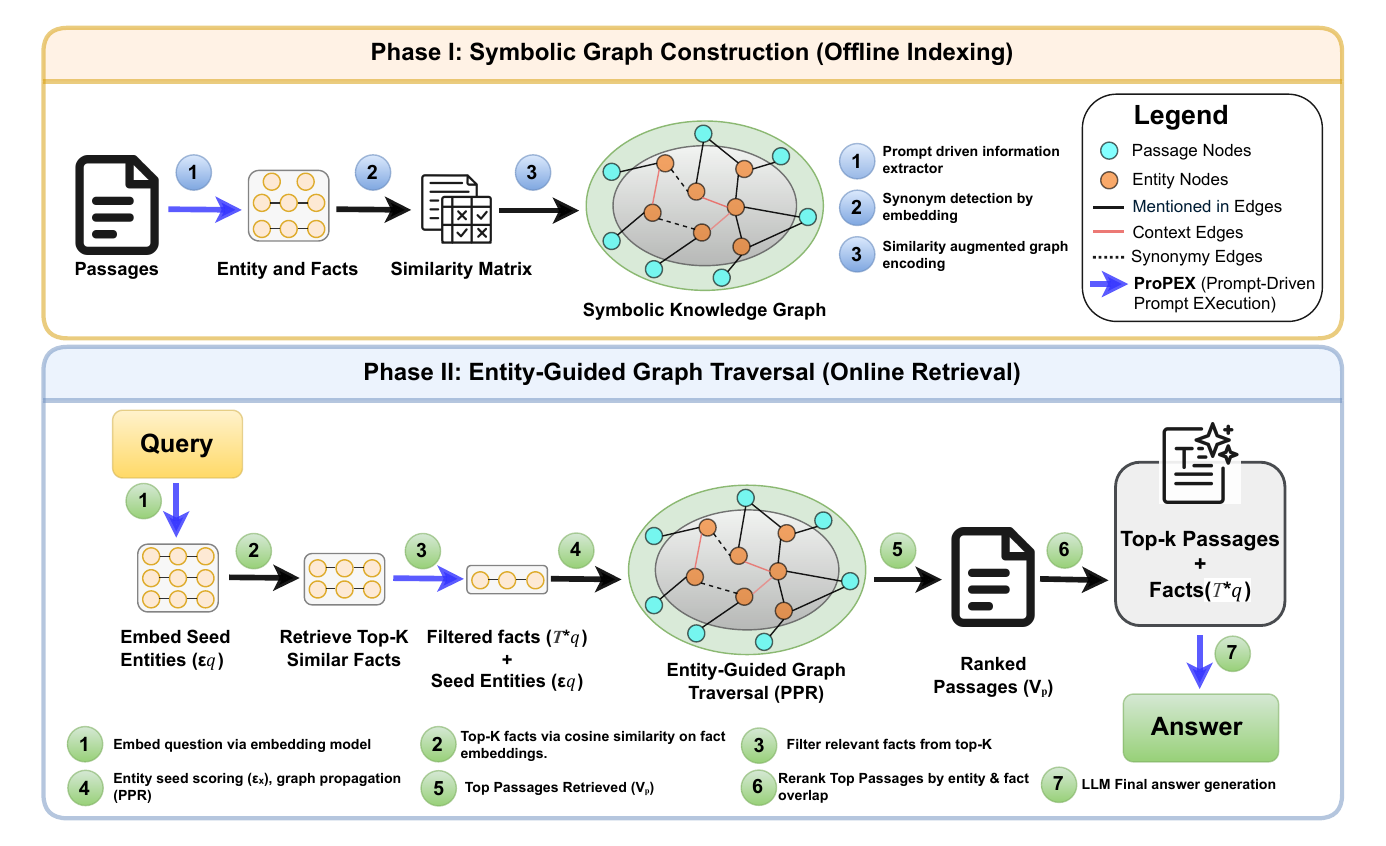}
\caption{Architecture of our retrieval-augmented QA framework. Phase I constructs a symbolic knowledge graph from LLM-extracted entities and facts triples. Phase II performs PPR-based traversal using query-aligned seeds and filtered facts to retrieve and re-rank passages for grounded answer generation.}
\label{fig:architecture}
\end{figure}

We propose a prompt-driven multi-hop RAG framework built around a novel Prompt-Driven Prompt Execution (ProPEX) mechanism that actively guides the retrieval process. Our system constructs a symbolic entity-centric knowledge graph from LLM-extracted factual triples and employs prompt-conditioned LLM filtering to select relevant facts. Inference, query prompts drive both seed entity selection and a Personalized PageRank (PPR) traversal over the graph to retrieve contextually linked passages. This integration of prompt design with structured graph reasoning enables interpretable, scalable, and accurate multi-hop QA.  The overall methodology is illustrated in {\hypersetup{linkcolor=softblue}\hyperref[fig:architecture]{Figure~\ref*{fig:architecture}}}.

\subsection{Symbolic Graph Construction (Offline Indexing)}
To enable efficient and interpretable multi-hop reasoning, we implement a symbolic knowledge graph as a persistent memory layer. A modular, prompt-guided pipeline extracts high-precision entities and factual assertions using a two-stage, LLM-assisted process. The prompts identify salient entities such as persons, organizations and locations, forming factual triples \( T = \{(s, p, o)\} \) where the subject \( s \) and the object \( o \) are structurally coherent entities. In a knowledge graph, the entity and the passage nodes are connected via directed mentioned-in links, forming typed relational edges within a heterogeneous graph. This structure includes bidirectional synonymy edges (from dense embedding similarity) and contextual relatedness edges derived from \( T \) co-occurrence data. Node scores are computed using the inverse passage frequency, while a sparse entity-to-passage incidence matrix enables retroactive reasoning and localized exploration. The resulting symbolic graph provides a query-time index for controlled, multi-hop retrieval and inference.

\vspace{-12pt} 

\subsection{Entity-Guided Graph Traversal (Online Retrieval)}
\noindent
At inference time, we perform entity-centric multi-hop retrieval using a symbolic propagation framework based on PPR over a heterogeneous knowledge graph $\mathcal{G} = (\mathcal{V}, \mathcal{E})$, where nodes represent entities or passages, and edges encode typed semantic relations. Given a query $q$, we compute its embedding $\mathbf{q}$ and retrieve top-$k$ fact triples $\mathcal{T}_q = \{(s_i, p_i, o_i)\}$ via cosine similarity with pre-embedded triples. These are refined by an LLM to obtain $\mathcal{T}_q^*$, from which we extract seed entities:
\begin{equation}
\mathcal{E}_q = \{ s_i, o_i \mid (s_i, p_i, o_i) \in \mathcal{T}_q^* \}
\end{equation}

We initialize a restart distribution $\mathbf{v}^{(0)}$ over seed entities:
\begin{equation}
\mathbf{v}_e^{(0)} = 
\begin{cases}
    \frac{\alpha}{|\mathcal{E}_q|}, & \text{if } e \in \mathcal{E}_q \\
    0, & \text{otherwise}
\end{cases}
\end{equation}
where $\alpha = 1.0$ controls the probability of restart. This is propagated across $\mathcal{G}$ through weighted edges representing \textbf{synonymy} ($w_\mathrm{sim}$), \textbf{contextual relatedness} ($w_\mathrm{rel}$), and references to the passage, using the iterative update of PPR:
\begin{equation}
\mathbf{v}^{(t+1)} = (1 - \alpha) \cdot \mathbf{A}^\top \mathbf{v}^{(t)} + \alpha \cdot \mathbf{v}^{(0)}
\end{equation}
where $\mathbf{A}$ is the normalized adjacency matrix. After convergence, the passage nodes accumulate relevance scores $\mathbf{v}_p$ that reflect their multi-hop connectivity to the query.

The final retrieval is performed by re-ranking:
\begin{equation}
\text{Retrieve}(q) = \operatorname{TopK}\left(\mathbf{v}_p + \lambda_\mathrm{rerank} \cdot \text{overlap}(p, q)\right)
\end{equation}
where $\text{overlap}(p, q)$ rewards symbolic alignment with query entities and filtered facts. This PPR-guided traversal enables interpretable, prompt-conditioned reasoning for structured multi-hop QA tasks.

\subsubsection{Answer Generation:}
\noindent
Following PPR-based symbolic propagation, the top$k$ passages reflecting multi-hop relational alignment within the knowledge graph are selected as the evidence context. To synthesize the final answer, we construct a structured prompt containing: (i) the ranked passages, (ii) seed entities $\mathcal{E}_q$, (iii) LLM-filtered factual triples $\mathcal{T}_q^*$, and (iv) a directive to produce an evidence-based response. A deterministic decoding strategy is employed using GPT-4.1 mini in zero- or few-shot mode without fine-tuning. This prompt-executed generation enhances factual consistency, reduces hallucinations, and ensures traceable, retrieval-aligned answers.

\section{Results \& discussion}
\subsection{Baselines}
We benchmark our framework against representative dense, multi-hop, and symbolic retrieval baselines. Dense retrievers include DPR~\cite{karpukhin2020dense} and RAG~\cite{lewis2020retrieval}, which retrieve passages through embedding similarity and fuse them with generative models, we consider Contriever~\cite{izacard2021unsupervised} and GTR~\cite{muennighoff2024generative}, both widely used embedding-based retrievers. From the category of large embedding models, we include GritLM/GritLM-7B~\cite{ni2021large} and NV-Embed-v2~\cite{lee2024nv}. For structure-augmented methods, we benchmark against RAPTOR~\cite{sarthi2024raptor}, which organizes the corpus hierarchically by semantic similarity, GraphRAG~\cite{edge2024local} and LightRAG~\cite{guo2024lightrag}, both of which employ knowledge graph structures for reasoning, and HippoRAG2~\cite{gutierrez2025rag}, a recent extension that integrates entity-guided Personalized PageRank for multi-hop retrieval.

\subsection{Implementation Details}
\noindent
Our framework is training-free and lightweight. Symbolic memory is constructed once per corpus and is accessed read-only at inference. At query time, the controller applies a seed policy, a compact keep–drop prompt for fact gating, and a typed PPR traversal with fixed damping. Entity mentions are extracted with GPT-4.1-mini, and similarity is scored using text-embedding-3-large. For each query, the top-5 triples are retained, and the QA module conditions on these passages with an evidence-first prompt under deterministic decoding.

\vspace{-12pt} 
\begin{table}[htb]
    \centering
    \caption{Statistics of the multi-hop QA datasets used in our evaluation.}
    \label{tab:dataset_stats}
    \begin{tabular}{|l|c|c|}
        \hline
        \textbf{Dataset} & \textbf{Number of Queries} & \textbf{Number of Passages} \\
        \hline
        HotpotQA & 1,000 & 9,811 \\
        2WikiMultihopQA & 1,000 & 6,119 \\
        \hline
    \end{tabular}
\end{table}
\vspace{-20pt} 

\subsection{Dataset Description}
\noindent
We evaluated our system using two prominent multi-hop question answering datasets: \textbf{HotpotQA}~\cite{yang2018hotpotqa} and \textbf{2WikiMultihopQA}~\cite{ho2020constructing}. These datasets are designed to assess complex reasoning across texts, making them suitable for measuring retrieval and inference efficiency in symbolic graph-enhanced RAG systems. HotpotQA focuses on linking facts with confirmed supporting information, while 2WikiMultihopQA presents a wider range of evidential diversity and challenges of associative retrieval. {\hypersetup{linkcolor=softblue}\hyperref[tab:dataset_stats]{Table~\ref*{tab:dataset_stats}}} provides an overview of the statistics of the data set.

\begin{table}[t!]
\caption{QA performance Exact Match(EM) and F1 scores on 2WikiMultiHopQA and HotpotQA datasets. Bold values indicate the best scores}
\label{tab:qa_results}
\begin{tabular*}{\textwidth}{@{\extracolsep\fill}lcccccc}
\toprule
\multirow{2}{*}{\textbf{Method}} & \multicolumn{2}{@{}c@{}}{\textbf{2WikiMultiHopQA}} & \multicolumn{2}{@{}c@{}}{\textbf{HotpotQA}} & \multicolumn{2}{@{}c@{}}{\textbf{Average}} \\
\cmidrule{2-3} \cmidrule{4-5} \cmidrule{6-7}
 & \textbf{EM} & \textbf{F1} & \textbf{EM} & \textbf{F1} & \textbf{EM} & \textbf{F1} \\
\midrule
\rowcolor{gray!20}
\multicolumn{7}{c}{\textbf{\textit{Simple Baselines}}} \\
Contriever~\cite{izacard2021unsupervised}  & 38.1 & 41.9 & 51.3 & 62.3 & 44.7 & 52.1 \\
GTR (T5-base)~\cite{muennighoff2024generative} & 49.2 & 52.8 & 50.6 & 62.8 & 49.9 & 57.8 \\
\rowcolor{gray!20}
\multicolumn{7}{c}{\textbf{\textit{Large Embedding Models}}} \\
GritLM-7B~\cite{ni2021large}  & 55.8 & 60.6 & 60.7 & 73.3 & 58.2 & 66.9 \\
NV-Embed-v2 (7B)~\cite{lee2024nv}  & 57.5 & 61.5 & 62.8 & 75.3 & 60.1 & 68.4 \\
\rowcolor{gray!20}
\multicolumn{7}{c}{\textbf{\textit{Structure-Augmented RAG}}} \\
RAPTOR~\cite{sarthi2024raptor}          & 47.3 & 52.1 &  56.8 &  69.5 & 52.0 & 60.8 \\
GraphRAG~\cite{edge2024local}           & 51.4 & 58.6 & 55.2 & 68.6 & 53.3 & 63.6 \\
LightRAG~\cite{guo2024lightrag}         & 9.4 & 11.6 & 2.0 & 2.4 & 5.7 & 7.0 \\
HiRAG~\cite{huang2025retrieval}         & 69.0 & 74.4 & 62.0 & 72.9 & 65.5 & 73.6 \\
HippoRAG~\cite{jimenez2024hipporag}     & 65.0 & 71.8 & 52.6 & 63.5 & 58.8 & 67.6 \\
HippoRAG 2~\cite{gutierrez2025rag}      & 65.0 & 71.0 & 62.7 & 75.5 & 63.8 & 73.2 \\
\midrule
\textbf{ProPEX-RAG}            & \textbf{76.4} & \textbf{78.9} & \textbf{79.9} & \textbf{80.7} & \textbf{78.1} & \textbf{79.8} \\
\bottomrule
\end{tabular*}
\end{table}

\begin{table}[t!]
\caption{Retrieval performance (passage recall@5) on 2WikiMultiHopQA and HotpotQA datasets. Results show the percentage of queries with at least one gold-supporting passage in the top-5 retrieved candidates. GraphRAG, LightRAG and HiRAG are not presented because they do not directly produce passage retrieval results.}
\label{tab:retrieval_results}
\begin{tabular*}{\textwidth}{@{\extracolsep\fill}lccc}
\toprule
\textbf{Method} & \textbf{2WikiMultiHopQA} & \textbf{HotpotQA} & \textbf{Average} \\
\midrule
\rowcolor{gray!20}
\multicolumn{4}{c}{\textbf{\textit{Simple Baselines}}} \\
Contriever~\cite{izacard2021unsupervised}  &  57.5 & 75.3 & 66.4 \\
GTR (T5-base)~\cite{muennighoff2024generative} & 67.9 & 73.9 & 70.9 \\
\rowcolor{gray!20}
\multicolumn{4}{c}{\textbf{\textit{Large Embedding Models}}} \\
GritLM-7B~\cite{ni2021large}  & 76.0 & 92.4 & 84.2 \\
NV-Embed-v2 (7B)~\cite{lee2024nv}  & 76.5 & 94.5 & 85.5 \\
\rowcolor{gray!20}
\multicolumn{4}{c}{\textbf{\textit{Structure-Augmented RAG}}} \\
RAPTOR~\cite{sarthi2024raptor}       & 66.2 & 86.9 & 76.5 \\
HippoRAG~\cite{jimenez2024hipporag}  & 89.1 & 77.7 & 83.4 \\
HippoRAG 2~\cite{gutierrez2025rag}   & 90.4 & 96.3 & 93.3 \\
\midrule
\textbf{ProPEX-RAG}                        & \textbf{98.1} & \textbf{97.1} & \textbf{97.6} \\
\bottomrule
\end{tabular*}
\footnotetext{Note: All recall@5 values represent the percentage of queries where at least one gold-supporting passage appeared in the top-5 retrieved documents.}
\footnotetext[1]{2WikiMultiHopQA: A dataset for evaluating complex multi-hop question answering.}
\footnotetext[2]{HotpotQA: A dataset that challenges models with multi-hop reasoning across multiple passages.}
\end{table}

\subsubsection{Evaluation Metrics:}
\noindent
We evaluate on HotpotQA~\cite{yang2018hotpotqa} and 2WikiMultiHopQA~\cite{ho2020constructing} using Exact Match (EM) and token-level F1 for answer accuracy, and Recall@$k$ ($k \in \{1,2,5,8,10\}$) for retrieval quality. Recall@5 highlights early evidence grounding. All metrics are computed on held-out sets with per-query logging for detailed analysis.

\justifying
\subsubsection{Results:}
In this section we present the experimental results on two standard multi-hop QA benchmarks, HotpotQA~\cite{yang2018hotpotqa} and 2WikiMultihopQA~\cite{ho2020constructing} which are designed to test multi-step reasoning over diverse and compositional queries. To assess both retrieval and answer generation quality, we use the GPT-4.1 mini model as the decoding engine, conditioned on top-ranked passages retrieved through symbolic graph traversal. Our system demonstrates consistently strong performance across both datasets. The symbolic propagation mechanism, guided by entity-centric cues and fact-level context, enables a high recall of relevant passages within a restricted top-$k$ budget. When combined with our fact-aligned reranking strategy, the approach achieves robust end-to-end answer accuracy, outperforming many recent structure-aware RAG systems. Detailed results, including F1 and exact match metrics, together with retrieval recall comparisons, are summarised in 
{\hypersetup{linkcolor=softblue}\hyperref[tab:qa_results]{Table~\ref*{tab:qa_results}}}
and
{\hypersetup{linkcolor=softblue}\hyperref[tab:retrieval_results]{Table~\ref*{tab:retrieval_results}}}

\section{Conclusion}
\noindent
\justifying
We introduced ProPEX-RAG, an interpretable and prompt-driven QA framework that combines LLM-guided fact selection with Personalized PageRank-based traversal over symbolic knowledge graphs. The model beats the previously achieved SOTA. Further testing is being conducted on multiple other multi-hop datasets, and a second consideration is to assess how ProPEX-RAG handles scaling of corpus sizes.  

\section*{Acknowledgements}
We gratefully acknowledge the Walmart Center for Technical Excellence at IIT Madras for their generous support and resources that enabled this research.

%
%
%
%

\end{document}